\documentclass[lettersize,journal]{IEEEtran}
\usepackage{amsmath,amsfonts}
\usepackage{algorithmic}
\usepackage{algorithm}
\usepackage{array}
\usepackage[caption=false,font=normalsize,labelfont=sf,textfont=sf]{subfig}
\usepackage{textcomp}
\usepackage{stfloats}
\usepackage{url}
\usepackage{verbatim}
\usepackage{booktabs}
\usepackage{amsmath}
\usepackage{natbib}
\usepackage{graphicx}
\usepackage{cite}
\usepackage{xcolor}
\usepackage{soul}

\begin{document}

\title{Predict, Cluster, Refine: A Joint Embedding Predictive Self-Supervised Framework for Graph Representation Learning}

\author{
    Srinitish Srinivasan, Omkumar CU \\
    \IEEEauthorblockA{Vellore Institute of Technology, India \\ 
    Email: srinitish.srinivasan2021@vitstudent.ac.in, omkumar.cu@vit.ac.in}
}
\markboth{}%
{Shell \MakeLowercase{\textit{Srinitish et al.}}: Predict, Cluster, Refine: A Self-Supervised Framework for Graph Representation Learning}

\maketitle

\begin{abstract}
Graph representation learning has emerged as a cornerstone for tasks like node classification and link prediction, yet prevailing self-supervised learning (SSL) methods face challenges such as computational inefficiency, reliance on contrastive objectives, and representation collapse. Existing approaches often depend on feature reconstruction, negative sampling, or complex decoders, which introduce training overhead and hinder generalization. Further, current techniques which address such limitations fail to account for the contribution of node embeddings to a certain prediction in the absence of labeled nodes. To address these limitations, we propose a novel joint embedding predictive framework for graph SSL that eliminates contrastive objectives and negative sampling while preserving semantic and structural information. Additionally, we introduce a semantic-aware objective term that incorporates pseudo-labels derived from Gaussian Mixture Models (GMMs), enhancing node discriminability by evaluating latent feature contributions. Extensive experiments demonstrate that our framework outperforms state-of-the-art graph SSL methods across benchmarks, achieving superior performance without contrastive loss or complex decoders. Key innovations include (1) a non-contrastive, view-invariant joint embedding predictive architecture, (2) leveraging single context and multiple targets relationship between subgraphs, and (3) GMM-based pseudo-label scoring to capture semantic contributions. This work advances graph SSL by offering a computationally efficient, collapse-resistant paradigm that bridges spatial and semantic graph features for downstream tasks.
\end{abstract}


\section{Introduction}
\label{sec:introduction}
Graph representation learning has become a cornerstone of modern machine learning, powering advancements in domains as diverse as Social Network Analysis, Recommendation Systems, and Drug Discovery~\citep{wu2020comprehensive}. The core promise is to distill complex, high-dimensional graph structures into potent, low-dimensional embeddings that capture essential spatial and spectral features. These embeddings can then fuel downstream tasks like node classification~\citep{maekawa2022beyond} and link prediction~\citep{zhang2018link} with remarkable efficiency, often by training simple decoders on a fixed, pre-trained backbone. However, the dominant paradigm, Graph Neural Networks (GNNs), faces a critical bottleneck: their performance is fundamentally tied to the availability of large, labeled datasets, which are often scarce and expensive to acquire. This dependency severely limits their applicability in real-world, data-sparse scenarios.

To break free from this supervised learning constraint, Graph Self-Supervised Learning (G-SSL) has emerged as a powerful alternative. Pioneers like Node2Vec, DGI~\citep{velivckovic2018deep}, and GRACE~\citep{zhu2020deep} have demonstrated impressive results by learning from the graph's intrinsic structure. Yet, a closer inspection reveals that the current G-SSL landscape is dominated by two problematic paradigms. The first, contrastive learning, relies on manually-defined graph augmentations and the generation of negative samples to pull similar nodes together and push dissimilar ones apart. This process is not only computationally prohibitive on large-scale graphs but is also highly sensitive to the choice of views and negative sampling strategy, often introducing noise and instability. The second, generative learning, focuses on reconstructing masked features or graph structures. These methods are notoriously fragile, often depending on specific decoder architectures and requiring complex designs like skip connections to combat vanishing gradients. Furthermore, both paradigms frequently resort to deep, multi-layer encoders to capture long-range dependencies, inadvertently falling prey to the pervasive issue of over-smoothing, which erodes node distinguishability and cripples downstream performance~\citep{WANG2024106484}.

To transcend these limitations, we introduce a fundamentally different approach: a joint embedding predictive framework that sidesteps the pitfalls of both contrastive and generative methods. Our model learns by predicting the latent representations of a target subgraph from the representation of a context subgraph, conditioned on a shared latent variable. This is achieved through an elegant asymmetric architecture: a context encoder, trained via standard gradient descent on augmented graph views, and a target encoder, which generates stable, high-quality representations from the original graph and is updated as an exponential moving average of the context encoder. This design completely obviates the need for negative sampling and fragile reconstruction objectives

Crucially, our framework incorporates two novel mechanisms to ensure rich and well-distributed embeddings. First, to combat representational collapse, we predict multiple, randomly masked target subgraph embeddings from a single context, forcing the model to learn a more diverse and informative latent space. This is further enhanced by injecting the positional information of target nodes into the context. Second, to ensure the embeddings are not just discriminative but also semantically coherent, we introduce a powerful regularization term. We fit a Gaussian Mixture Model (GMM) on the learned embeddings to generate pseudo-labels and then score the contribution of each latent feature to these clusters. This forces the model to capture high-level community structure within the graph. The result is a robust, augmentation-invariant G-SSL technique that learns powerful representations without the computational overhead and design complexities of its predecessors.

We list our main contributions as follows:
\begin{itemize}
    \item We propose a novel joint embedding predictive framework that obviates the need for contrastive objectives, negative sampling, and explicit graph reconstruction, offering a more efficient and direct learning paradigm.
    
    \item We introduce a multi-target prediction strategy coupled with positional encoding to explicitly combat representational collapse, ensuring a well-distributed and informative embedding space.
    
    \item We pioneer a semantic regularization objective that scores the contribution of learned embeddings to GMM-derived pseudo-labels, guiding the model to capture meaningful high-level community structures.
    
    \item We conduct extensive experiments demonstrating that our approach significantly outperforms previous state-of-the-art G-SSL methods and provide a thorough empirical analysis highlighting its superior efficiency and scalability.
\end{itemize}

\section{Related Work}
\subsection{Unsupervised Representation Learning on Graphs}
\citep{jin2021automated} proposes a method for composing multiple self-supervised tasks for GNNs. They introduce a pseudo-homophily measure to evaluate representation quality without labels. \citep{zhang2021canonical} proposed the removal of negative sampling and MI estimator optimization entirely. The authors propose a loss function that contains an invariance term that maximizes correlation between embeddings of the two views and a decorrelation term that pushes different feature dimensions to be uncorrelated. \citep{hou2022graphmae} employs a re-mask decoding strategy and uses expressive GNN decoders instead of Multi-Layer Perceptrons(MLPs) enabling the model to learn more meaningful compressed representations. It lays focus on masked feature reconstruction rather than structural reconstruction. \citep{hassani2020contrastive} proposed MVGRL that makes use of two graph views and a discriminator to maximize mutual information between node embeddings from a first-order view and graph embeddings from a second-order view. It leverages both node and graph-level embeddings and avoids the need for explicit negative sampling. \citep{ju2022multi} proposed ParetoGNN which simultaneously learns from multiple pretext tasks spanning different philosophies. It uses a multiple gradient descent algorithm to dynamically reconcile conflicting learning objectives, showing state-of-the-art performance in node classification, clustering, link prediction and community prediction. 

\subsection{Bootstrapping Methods}
 \citep{ding2023eliciting} introduced  multi-scale feature propagation to capture long-range node interactions without oversmoothing. The authors also enhance inter-cluster separability and intra-cluster compactness by inferring cluster prototypes using a Bayesian non-parametric approach via Dirichlet Process Mixture Models(DPMMs). \citep{thakoor2021large} makes use of simple graph augmentations such as random node feature and edge masking, making it easier to implement on large graphs while achieving state-of-the-art results. It leverages a cosine similarity-based objective to make the predicted target representations closer to the true representations. \citep{jin2021multi} introduced a Siamese network architecture comprising an online and target encoder with momentum-driven update steps for the target. The authors propose 2 contrastive objectives i.e cross-network and cross view-contrastiveness. The Cross-network contrastive objective incorporates negative samples to push disparate nodes away in different graph views to effectively learn topological information.

\subsection{Joint Predictive Embedding Methods}
Joint predictive embedding has been explored in the field of computer vision and audio recognition. \citep{assran2023self} introduced I-JEPA which eliminates the need for hand-crafted augmentations and image reconstruction. They make use of a joint-embedding predictive model that predicts representations of masked image regions in an abstract feature space rather than in pixel space, allowing the model to focus on high-level semantic structures. It makes use of a Vision Transformer(ViT) with a multi-block masking strategy ensuring that the predictions retain semantic integrity. \citep{fei2023jepa}  extends the masked-modeling principle from vision to audio, enabling self-supervised learning on spectrograms. The key technical contribution is the introduction of a curriculum masking strategy, which transitions from random block masking to time-frequency-aware masking, addressing the strong correlations in audio spectrograms.

\section{Methodology}

\begin{figure*}[h!]
    \centering
    \includegraphics[width=\linewidth]{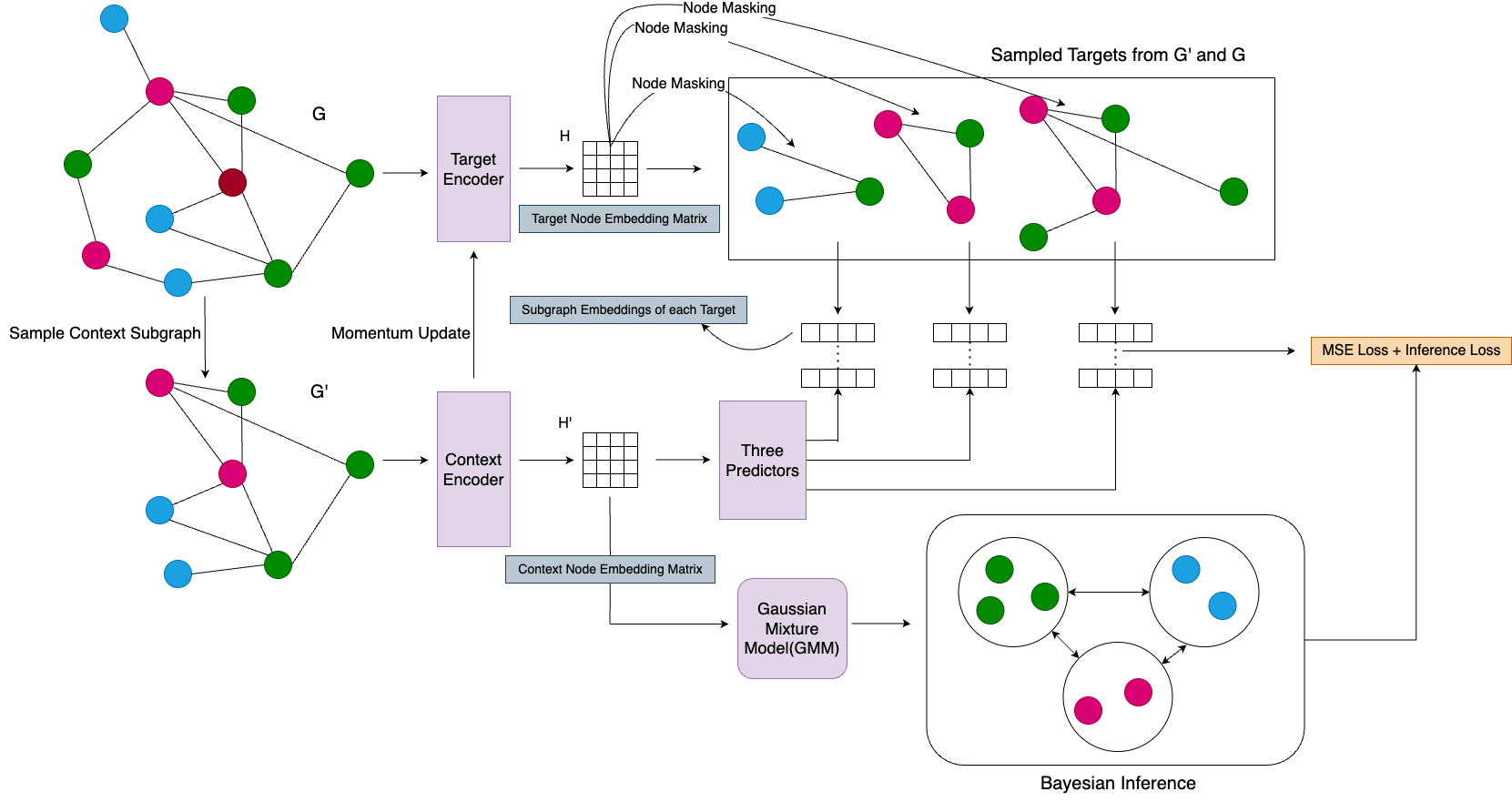}
    \caption{The Proposed Self-Supervised Learning Architecture. The framework operates in two parallel streams. The primary predictive task involves an asymmetric encoder design. A Context Encoder processes an augmented graph view to produce a context embedding. This is used to predict the representations of multiple, distinct Target Subgraphs, which are generated by a momentum-based Target Encoder from the original graph. This multi-target prediction prevents the model from learning trivial solutions. A Semantic Consistency Module runs in parallel, fitting a Gaussian Mixture Model (GMM) on the learned node embeddings to derive pseudo-labels. These labels are then used to compute a regularization term that re-weights the embeddings, ensuring they capture coherent, high-level graph structures. The final objective is a weighted sum of the predictive loss and the semantic consistency score.}
    \label{fig:diag}
\end{figure*}
\subsection{Preliminaries}
Consider the definition of a Graph $G=(V,E)$. Let $V$ be the set of vertices $\{v_1,v_2,v_3....v_{n_v}\}$ and $E$ be the set of edges $\{e_1,e_2,e_3...e_{n_e}\}$. $n_v,n_e$ are
the number of nodes and edges respectively in $G$. Each node in $V$ is characterized by a $d$ dimensional vector. This is the initial signal that is populated by either bag of words or binary values depending on the problem considered.

\subsection{Loss Function}
\subsubsection{Joint Predictive Optimization}
We sample a subgraph $G'$ from graph $G$ by dropping a set of nodes according to a Bernoulli distribution parameterized by success probability $p_1$. The subgraph $G'$ is passed into the context encoder to output node embeddings $H'$ of dimensions $d'$. Meanwhile, the graph $G$ is passed into the target encoder to generate node embeddings $H$ with the same dimensions as $H'$. At the latent space, we sample 3 target subgraphs from the context subgraph $G'$, again according to a Bernoulli distribution with probability $p_2$ such that $p_2<p_1$. The node embeddings of each subgraph are obtained by deactivating the node features of masked nodes, which is followed by global mean pooling operation, thus obtaining embeddings $H^t_1$,$H^t_2$ and $H^t_3$. The context node level embeddings $H'$ is passed into 3 predictors, corresponding to the three target subgraphs. The obtained predictions are then pooled by the mean pooling operation, thus resulting in target predictions $H'_1,H'_2,H'_3$, conditioned on latent variable $H'$. We then frame the objective function for the joint predictive component as follows,

\begin{IEEEeqnarray}{rCl}
    \mathcal{L}^J(\Theta^C,\Theta^R) &=& \frac{1}{T} \sum_{k=1}^{T} \| H'_k - H^t_k \|^2
\end{IEEEeqnarray}
where $T$ is the total number of targets, $\Theta^C$ is the weight matrix of the context encoder, and $\Theta^R$ is the weight matrix of the target encoder. The objective is computed at the latent space with no reconstruction/negative sampling involved.  

\subsubsection{Node Feature Contribution Optimization}
The node embeddings $H'$ obtained from passing $G'$ into the context encoder are fit in a Gaussian Mixture Model (GMM). We aim to obtain:  
\begin{IEEEeqnarray}{rCl}  
    p(z_k=1|h'_n) &=& \frac{p(h'_n|z_k=1) p(z_k=1)}{\sum_{j=1}^{K} p(h'_n|z_j=1) p(z_j=1)}  
    \label{eq:gmm_eqn}  
\end{IEEEeqnarray}  
where $z$ is a latent variable that takes two values: one if $h'$ comes from Gaussian $k$, and zero otherwise. We can obtain:  
\begin{IEEEeqnarray}{rCl}  
    p(z_k=1) &=& \pi_k  \\  
    p(h'_n|z_k=1) &=& \mathcal{N}(h'_n|\mu_k,\Sigma_k)  
\end{IEEEeqnarray}  
where $\pi_k$ is the mixing coefficient such that:  
\begin{IEEEeqnarray}{rCl}  
    \sum_{k=1}^{K} \pi_k &=& 1  
\end{IEEEeqnarray}  

By substituting these equations into \eqref{eq:gmm_eqn}, we obtain:  
\begin{IEEEeqnarray}{rCl}  
    p(z_k=1|h'_n) &=& \frac{\pi_k \mathcal{N}(h'_n|\mu_k,\Sigma_k)}{\sum_{j=1}^{K} \pi_j \mathcal{N}(h'_n|\mu_j,\Sigma_j)}  
    = \gamma(z_{nk})  
    \label{eqn:gamma}  
\end{IEEEeqnarray}  

\textbf{E-Step:}  
In the E-Step, we aim to evaluate:  
\begin{IEEEeqnarray}{rCl}  
    \mathcal{Q}(\theta^*,\theta) &=& \mathbb{E}[\ln p(H',Z|\theta^*)] \nonumber \\  
    &=& \sum_{Z} p(Z|H',\theta) \ln p(H',Z|\theta^*)  
\end{IEEEeqnarray}  
From \eqref{eqn:gamma}, we substitute $\gamma$ into the equation:  
\begin{IEEEeqnarray}{rCl}  
    \mathcal{Q}(\theta^*,\theta) &=& \sum_{Z} \gamma(z_{nk}) \ln p(H',Z|\theta^*)  
\end{IEEEeqnarray}  
The complete likelihood of the model then becomes:  
\begin{IEEEeqnarray}{rCl}  
    \mathcal{Q}(\theta^*,\theta) &=& \sum_{n=1}^{N} \sum_{k=1}^{K} \gamma(z_{nk}) \left[ \ln \pi_k + \ln \mathcal{N}(h'_n|\mu_k,\Sigma_k) \right]  
    \label{eq:e_step}  
\end{IEEEeqnarray}  

\textbf{M-Step:}  
In the M-Step, we aim to find updated parameters $\theta^*$ as follows:  
\begin{IEEEeqnarray}{rCl}  
    \theta^* &=& \arg\max_\theta \mathcal{Q}(\theta^*,\theta)  
\end{IEEEeqnarray}  
Considering the constraint $\sum_{k=1}^{K} \pi_k = 1$, equation \eqref{eq:e_step} is modified as follows:  
\begin{IEEEeqnarray}{rCl}  
    \mathcal{Q}(\theta^*,\theta) &=& \sum_{n=1}^{N} \sum_{k=1}^{K} \gamma(z_{nk}) \Big[ \ln \pi_k + \ln \mathcal{N}(h'_n|\mu_k,\Sigma_k) \Big]  \nonumber \\  
    && - \lambda \left(\sum_{k=1}^{K} \pi_k - 1 \right)  
\end{IEEEeqnarray}  

The parameters are then determined by maximizing $\mathcal{Q}$. Taking the derivative with respect to $\pi_k, \mu_k, \Sigma_k$ and rearranging terms, we obtain the update equations:  
\begin{IEEEeqnarray}{rCl}  
    \pi_k &=& \frac{\sum_{n=1}^{N} \gamma(z_{nk})}{N}  
\end{IEEEeqnarray}  
\begin{IEEEeqnarray}{rCl}  
    \mu_k^* &=& \frac{\sum_{n=1}^{N} \gamma(z_{nk}) h'_n}{\sum_{n=1}^{N} \gamma(z_{nk})}  
\end{IEEEeqnarray}  
\begin{IEEEeqnarray}{rCl}  
    \Sigma_k^* &=& \frac{\sum_{n=1}^{N} \gamma(z_{nk}) (h'_n - \mu_k) (h'_n - \mu_k)^T}{\sum_{n=1}^{N} \gamma(z_{nk})}  
\end{IEEEeqnarray}  

\textbf{Parameter $\Theta^C$ Update:  }
Let $V_g$ be a vector of pseudo-labels obtained for each node from the Gaussian Mixture Model(GMM) and $V_k$ be the vector of pseudo-labels obtained by clustering node embeddings $H'$ by K-Means. We update the context encoder parameters $\Theta^c$ with the following objective, 
\begin{IEEEeqnarray}{rCl}
    \mathcal{L}^G &=&
    \begin{cases}
    \frac{1}{2} \left(\dfrac{V_g^T H'}{\sum V_g \|V_g\|} - \dfrac{V_k^T H'}{\sum V_k \|V_k\|} \right) / \beta, \\ 
    \quad \text{if} \ \left| \dfrac{V_g^T H'}{\sum V_g \|V_g\|} - \dfrac{V_k^T H'}{\sum V_k \|V_k\|} \right| < \beta, \\[10pt]
    
    \left| \dfrac{V_g^T H'}{\sum V_g \|V_g\|} - \dfrac{V_k^T H'}{\sum V_k \|V_k\|} \right| - \frac{1}{2} \beta, \\ 
    \quad \text{otherwise}.
    \end{cases}
\end{IEEEeqnarray}

In the above equation, the context encoder parameters $\Theta^c$ are updated by the smooth $L_1$ loss function.

\subsubsection{Final objective}
The context encoder parameters $\Theta^C$ are finally updated as follows,
\begin{equation}
    \mathcal{L}=\mathcal{L^J}+\mathcal{L^G}
\end{equation}
\begin{equation}
    \Theta^C\leftarrow\text{optimize}(\Theta^c,\alpha,\partial_{\Theta^c}\mathcal{L})
\end{equation}
where, $\alpha$ is the learning rate for the Adam optimizer. The weights of the context and target encoder are randomly initialized by a standard normal distribution. A cosine annealing learning rate scheduler with early stopping is used in all experiments.

\subsection{Description of Components and Operations}
\subsubsection{Context Encoder}
The node features from $G'$ are passed into the context encoder. The context encoder is a simple 3-layer GCN encoder which predicts 128, 256 and 512 features in each layer respectively. The forward propagation for each layer is described as follows,
\begin{align}
    X'=g(\hat{D}^{\frac{-1}{2}}\hat{A}\hat{D}^{\frac{=1}{2}}X\Theta)\\
    \hat{A}=A+I
\end{align}
where $\hat{D}$ is the degree matrix, $\hat{A}$ is the adjacency matrix with added self-loops and $\Theta$ is the learned weights matrix. $g$ is a non-linear function. We use ReLU for the first two layers and Tanh for the final layer. The context encoder is an online encoder whose weights are updated by gradient descent.

\subsubsection{Target Encoder}
The target encoder inputs node features from $G$. Similar to the context encoder, it is a 3-layer GCN encoder with same number of hidden, output dimensions and forward propagation steps. The weights of the target encoder are updated as a moving average of the context encoder as follows,
\begin{equation}
    \Theta_{s}^R=m\Theta^R_{s-1}+(1-m)\Theta_s^C
\end{equation}
where $\Theta^R$ is the weights matrix of the target encoder, $\Theta^C$ is the weights of the context encoder, $m$ is the momentum parameter and $s$ refers to the $s$th iteration.

\subsubsection{Predictor}
The predictor consists of 2 GCN layers, both predicting 512 features. The activation function $g$ for both layers is Tanh, in line with the final layer of the target encoder. We use Tanh since it is a bounded function, thus stabilizing the loss computation and optimization.

\subsubsection{Global Mean Pooling Operation}
The global mean pool operation for a graph $G$ is given as follows,
\begin{equation}
    r=\frac{1}{N}\Sigma_{n=1}^{n=N}x_n
\end{equation}
where $x_n$ refers to the node features of node $n$ and $N$ is the total number of nodes in graph $G$.

\subsection{Implementation Details}
For all experiments, we use the Adam Optimizer\citep{kingma2014adam} and cosine annealing learning rate scheduler with warm restarts. No regularization techniques such as dropout, $L_1$ or $L_2$ have been employed since they tend to reduce performance. For learning rates, we perform a search with values corresponding to the search space $\{0.5,0.1,0.05,0.01,0.001\}$. We set the value of the momentum update parameter $m$ to 0.9. The number of epochs was set to 50,000 with early stopping. All experiments were conducted on an M2 Macbook Pro with 8GB CPU.

\section{Experiments}
\subsection{Experimental Setting}
\subsubsection{Datasets}
In our experiments, we evaluate our proposed framework on seven publicly available benchmark datasets for node representation learning and classification. The datasets are, namely, Cora\citep{sen2008collective}, Pubmed\citep{namata2012query}, Citeseer\citep{sen2008collective}, Amazon Photos\citep{shchur2018pitfalls}, Amazon Computers\citep{shchur2018pitfalls}, Coauthor CS\citep{shchur2018pitfalls} and WikiCS\citep{mernyei2020wiki}. Details on number of nodes, edges and features are given in table \ref{tab:data}. Cora consists of 2708 scientific publications classified into one of seven classes, Citeseer consists of 3312 scientific publications classified into one of six classes and Pubmed consists of 19717 scientific publications pertaining to diabetes classified into one of three classes. In Amazon Photos and Amazon Computers, nodes represent goods and edges represent that two goods are frequently bought together. The product reviews are represented as bag-of-words features. Coauthor CS contains paper keywords for each author's papers. Nodes represent authors that are connected by an edge if they co-authored a paper. WikiCS is a dataset derived from Wikipedia with nodes corresponding to Computer Science articles and edges based on hyperlinks. The 10 classes represent different branches of Computer Science. 

\begin{table}[h!]
    \centering
    \caption{\textbf{Dataset Details}}
    \begin{tabular}{lcccccc}
        \toprule
        \textbf{Dataset} & \textbf{Nodes} & \textbf{Edges} & \textbf{Features} & \textbf{Classes} \\
        \midrule
        Cora             & 2,708          & 10,556         & 1,433             & 7                \\
        Citeseer         & 3,327          & 9,104          & 3,703             & 6                \\
        Pubmed           & 19,717         & 88,648         & 500               & 3                \\
        Amazon Photos    & 7,650          & 238,612        & 745               & 8                \\
        Amazon Computers & 13,752         & 491,722        & 767               & 10               \\
        Coauthor CS      & 18,333         & 163,788        & 6,805             & 15               \\
        WikiCS           &  11,701              & 216,123               &300                   & 10                 \\
        \bottomrule
    \end{tabular}
    \label{tab:data}
\end{table}

\subsubsection{Evaluation Methodology}
The embeddings are evaluated on node classification, whose performance is quantified by accuracy. We follow the same evaluation protocol as used in \citep{velivckovic2017graph}, \citep{ding2023eliciting}, \citep{ju2022multi} etc. All scores for baselines have been obtained from previously published papers. 

\subsubsection{Evaluation Protocol}
For all downstream tests, we follow the linear evaluation protocol on graphs where the parameters of the backbone/encoder are frozen during inference time. Only the prediction head, which is a single GCN layer, is trained for node classification. For evaluation purposes, we use the default number for train/val/test splits for the citation networks i.e. Cora, Pubmed, Citeseer which are 500 validation and 1000 testing nodes. The train/val/test splits for the remaining datasets, namely, Amazon Photos, Amazon Computers and Coauthor CS are according to \citep{shchur2018pitfalls}. For WikiCS we use the publicly available splits. Unless otherwise mentioned, performance is averaged over 10 independent runs with different random seeds and splits for all seven evaluation datasets. We report the mean and standard deviation obtained across 10 runs. 

\subsubsection{Environment}
To ensure a fair comparison, for all datasets, we use the same number of layers for the GNN encoder and the same number of features at each hidden layer. The encoder predicts 512 hidden features in the latent space for all datasets. The number of output features of the final classification head is the only parameter varied. We use a Cosine Annealing learning rate scheduler with warm restarts after 75 epochs, and early stopping is employed for all experiments. 

\subsection{Evaluation Results}
\subsubsection{Performance on Node Classification on small and large graphs}
As mentioned earlier, we use the linear evaluation protocol and report the mean and standard deviation of classification accuracy on the test nodes over 10 runs on different folds or splits with different seeds. We compare our proposed approach with supervised and fine-tuned model baselines. For semi-supervised node classification, we compare our proposed approach against Multi Layer Perceptron(MLP),Graph Convolution Network(GCN)\citep{kipf2016semi}, Graph Attention Network(GAT)\citep{velivckovic2017graph}, Simplified GCN\citep{wu2019simplifying}, Logistic Regression and GraphSAGE\citep{hamilton2017inductive}. For GraphSAGE, we use the mean, maxpool and meanpool variants as described in \citep{shchur2018pitfalls}. We have compared our model's performance against semi-supervised baselines in table \ref{tab:supervision}. For self-supervised and fine-tuned node classification, we compare our proposed approach against DGI, MVGRL\citep{hassani2020contrastive}, GRACE\citep{zhu2020deep}, CCA-SSG\citep{zhang2021canonical}, SUGRL\citep{mo2022simple}, S3-CL\citep{ding2023eliciting}, GraphMAE\citep{hou2022graphmae}, GMI\citep{peng2020graph}, BGRL\citep{thakoor2021large} and ParetoGNN\citep{ju2022multi} on small and large graphs.  In semi-supervised node classification (Table \ref{tab:supervision}), our method achieves the highest accuracy across all baselines. On self-supervised classification followed by fine-tuning, we evaluate our proposed model against two specific fields of data i.e Planetoid datasets and larger, more stable datasets such as Amazon Computers, Photos and WikiCS. Table \ref{tab:fine_tuned} compares our model's performance against strong baselines on Planetoid datasets. We show that our model performs extremely well on smaller datasets by outperforming all baselines on Cora and Citeseer with extremely competitive results on Pubmed. Table \ref{tab:med_datasets} contains the performance of our model against strong baselines for Photos, Computers, Coauthor CS and WikiCS. Our model consistently achieves competitive scores against baselines by outperforming baselines on Photos and CS while remaining competitive on Computers and WikiCS .These results underscore the approach's adaptability and strong performance across diverse dataset sizes. Owing to the ability of our proposed approach to avoid noisy features, representation collapse, and leverage semantic information, our proposed model performs well on small and unstable datasets such as Cora and Citeseer and large datasets such as Amazon, Coauthor and WikiCS. 

\begin{table*}[h!]
\centering
\footnotesize
\caption{\textbf{Semi-Supervised Node Classification.} The values indicate the accuracy achieved on node classification by several methods. A higher value indicates better performance. The best score is marked in \textbf{bold} and the second-best score is \underline{underlined}. N/A indicates that the score was not reported for a particular dataset by the original authors.}

\begin{tabular}{lccccccc}
\toprule

\textbf{Method}& \textbf{Cora} & \textbf{Citeseer} & \textbf{Pubmed} & \textbf{Photos} & \textbf{Computers} & \textbf{Coauthor CS} \\
\midrule
MLP & $55.2 \pm 0.4$ & $46.5 \pm 0.5$ & $71.4 \pm 0.3$ & $78.5 \pm 0.2$ & $44.9 \pm 5.8$ & $76.5 \pm 0.3$ \\
GCN & $81.5 \pm 1.3$ & \underline{$71.9 \pm 1.9$} & $77.8 \pm 2.9$ & $91.2 \pm 1.2$ & \underline{$82.6 \pm 2.4$} & $91.1 \pm 0.5$ \\
GAT & \underline{$81.8 \pm 0.3$} & $71.4 \pm 1.8$ & $78.7 \pm 2.3$ & $85.7 \pm 20.3$ & $78.0 \pm 19.0$ & $90.5 \pm 0.6$ \\
Simplified GCN & $81.5 \pm 0.2$ & $73.1 \pm 0.1$ & \underline{$79.7 \pm 0.4$} & $88.3 \pm 1.1$ & N/A & \underline{$91.5 \pm 0.3$} \\
GraphSage Mean & $79.2 \pm 7.7$ & $71.6 \pm 1.9$ & $77.4 \pm 2.2$ & \underline{$91.4 \pm 1.3$} & $82.4 \pm 1.8$ & $91.3 \pm 2.8$ \\
GraphSage MaxPool & $76.6 \pm 1.9$ & $67.5 \pm 2.3$ & $76.1 \pm 2.3$ & $90.4 \pm 1.3$ & N/A & $85.0 \pm 1.1$ \\
GraphSage MeanPool & $77.9 \pm 2.4$ & $68.6 \pm 2.4$ & $76.5 \pm 2.4$ & $90.7 \pm 1.6$ & $79.9 \pm 2.3$ & $89.6 \pm 0.9$ \\
Logistic Regression & $57.1 \pm 2.3$ & $61.0 \pm 2.2$ & $64.1 \pm 3.1$ & $73.0 \pm 6.5$ & $64.1 \pm 5.7$ & $86.4 \pm 0.9$ \\
Ours & \textbf{89.8 ± 0.9} & \textbf{77.0 ± 0.9} & \textbf{85.7 ± 1.0} & \textbf{94.5 ± 0.5} & \textbf{88.0 ± 0.6} & \textbf{94.5 ± 0.4} \\
\bottomrule
\end{tabular}
\label{tab:supervision}
\end{table*}

\begin{table*}[h!]
    \centering
    \caption{\textbf{Self Supervised Pre-Training followed by classification on Planetoid Datasets.} The values indicate the accuracy achieved on node classification by several SSL methods. A higher value indicates better performance. The best score is marked in bold and the second-best score is \underline{underlined}.}
    \renewcommand{\arraystretch}{1.2}
    \begin{tabular}{lccc}
        \toprule
       
       \textbf{Method} & \textbf{Cora} & \textbf{Citeseer} & \textbf{Pubmed} \\
        \midrule
        DGI             & $81.7\pm0.6$  & $71.5\pm0.7$  & $77.3\pm0.6$  \\
        MVGRL           & $82.9\pm0.7$  & $72.6\pm0.7$  & $79.4\pm0.3$  \\
        GRACE           & $80.0\pm0.4$  & $71.7\pm0.6$  & $79.5\pm1.1$  \\
        CCA-SSG         & $84.2\pm0.4$  & $73.1\pm0.3$  & $81.6\pm0.4$  \\
        SUGRL           & $83.4\pm0.5$  & $73.0\pm0.4$  & $81.9\pm0.3$  \\
        S3-CL           & $\underline{84.5 \pm 0.4}$ & $\underline{74.6 \pm 0.4}$ & $80.8 \pm 0.3$  \\
        GraphMAE        & $84.2\pm0.4$  & $73.1\pm0.4$  & $83.9\pm0.3$  \\
        GMI             & $82.7\pm0.2$  & $73.3\pm0.3$  & $77.3\pm0.6$  \\
        BGRL            & $83.8\pm1.6$  & $72.3\pm0.9$  & $\mathbf{86.0\pm0.3}$  \\
        Ours            & $\mathbf{89.8 \pm 0.9}$ & $\mathbf{77.0 \pm 0.9}$ & $\underline{85.7 \pm 1.0}$  \\
        \bottomrule
    \end{tabular}
    \label{tab:fine_tuned}
\end{table*}

\begin{table*}[h!]
    \centering
    \caption{\textbf{Self Supervised Pre-Training followed by classification on datasets mentioned in \citep{shchur2018pitfalls} and WikiCS.} The best score for each dataset is marked in bold and the second-best score is \underline{underlined}.}
    \renewcommand{\arraystretch}{1.2}
    \begin{tabular}{lcccc}
        \toprule
        \textbf{Method} & \textbf{Photos} & \textbf{Computers} & \textbf{Coauthor CS} & \textbf{WikiCS} \\
        \midrule
        DGI             & $91.6\pm0.2$  & $83.9\pm0.5$  & $92.1\pm0.6$  & $75.3\pm0.1$  \\
        GRACE           & $92.1 \pm 0.5$ & $86.7 \pm 0.8$ & $93.2 \pm 0.4$ & $77.5 \pm 0.6$ \\
        BGRL            & $93.2\pm0.3$  & \underline{$90.3\pm0.2$}  & \underline{$93.3\pm0.1$}  & $80.0\pm0.1$  \\
        ParetoGNN       & \underline{$93.8\pm0.3$}  & $\mathbf{90.7\pm0.2}$ & $92.2\pm0.1$  & $\mathbf{82.9\pm0.1}$ \\
        MVGRL           & $93.2\pm0.3$  & $87.5\pm0.1$  & $92.1\pm0.1$  & $77.5\pm0.1$  \\
        Random Weights  & $92.1\pm0.5$  & $86.5\pm0.4$  & $91.6\pm0.3$  & $78.9\pm0.6$  \\
        Ours            & $\mathbf{94.5 \pm 0.5}$ & $88.0 \pm 0.6$ & $\mathbf{94.5 \pm 0.4}$ &\underline{$82.4\pm1.0$}  \\
        \bottomrule
    \end{tabular}
    \label{tab:med_datasets}
\end{table*}

\subsubsection{Study on GMM Cluster Optimization}
In this section, we evaluate the significance of the constraint in our loss function. Our approach combines learning sub-graph embeddings with optimizing node embeddings using pseudo-labels derived from a Gaussian Mixture Model (GMM). These pseudo-labels, converted to normalized scores, serve as a constraint in the original loss function for subgraph embeddings. Table \ref{tab:ablation} illustrates the performance of our method with and without this constraint, demonstrating a substantial improvement when the GMM-derived pseudo-label scores are incorporated. This finding is particularly noteworthy as it indicates that computationally expensive processes like negative sampling, commonly used in graph contrastive learning, may not be essential for optimizing node embeddings. Our approach thus offers a more efficient alternative while maintaining, and even enhancing, performance in graph representation learning tasks.
\begin{table*}[h!]
    \centering
    \footnotesize
    \caption{\textbf{Ablation Study for Loss Function.} The best score for each dataset is marked in bold.}
    \begin{tabular}{lccccccc}
        \toprule
        \textbf{Method} & \textbf{Cora} & \textbf{Citeseer} & \textbf{Pubmed} & \textbf{Photos} & \textbf{Computers} & \textbf{Coauthor CS} &\textbf{WikiCS} \\
        \midrule
        Without Bayesian Inference & 89.0 & 74.1 & 84.2 & 93.4 & 85.0 & 93.2 &81.5 \\
        With Bayesian Inference    & \textbf{89.8} & \textbf{77.0} & \textbf{85.7} & \textbf{94.5} & \textbf{88.0} & \textbf{94.5} & \textbf{82.4}\\
        \bottomrule
    \end{tabular}
    \label{tab:ablation}
\end{table*}
\subsubsection{Study on Momentum Parameter $m$}
\begin{figure*}[h!]
    \centering
    \includegraphics[scale=0.7]{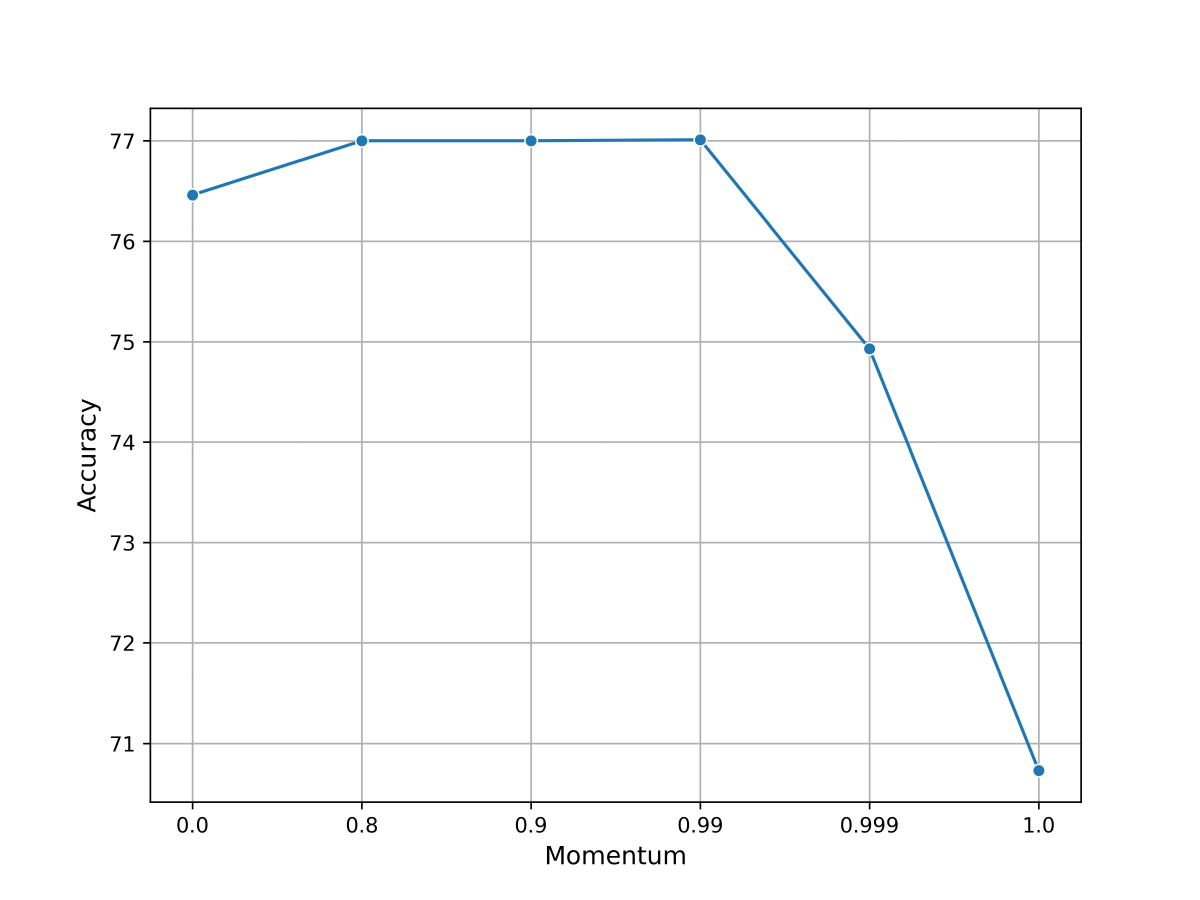}
    \caption{Variation of accuracy with respect to momentum parameter $m$ on Citeseer}
    \label{fig:var_plot}
\end{figure*}

Figure \ref{fig:var_plot} represents the change in performance with respect to the momentum parameter on the citeseer dataset. The ideal values for $m$ may range between 0.8 and 0.9. As $m\rightarrow 1$, the weights of the target encoder update by very small steps, thereby slowing down learning. This is further indicated by the drop in performance. For all our experiments, we use $m=0.9$.

\subsubsection{Efficiency Analysis for Scalability}
We evaluate the model's efficiency in terms of the parameter size and memory consumption. We show that our model is among the most efficient of all baselines while outperforming most baselines. We compare the efficiency of our proposed model against reported baselines in table \ref{tab:efficiency}. 

\begin{table*}[h!]
    \centering
    \caption{\textbf{Efficiency Comparison among Baselines on Cora, Pubmed, and Citeseer.} Lower the number of parameters, more efficient is the model}
    \begin{tabular}{lcc|cc|cc}
        \toprule
        \textbf{Methods} & \multicolumn{2}{c|}{\textbf{Cora}} & \multicolumn{2}{c|}{\textbf{Citeseer}} & \multicolumn{2}{c}{\textbf{Pubmed}} \\
        \cmidrule(lr){2-3} \cmidrule(lr){4-5} \cmidrule(lr){6-7}
        & \textbf{Memory (GB)} & \textbf{Params} 
        & \textbf{Memory (GB)} & \textbf{Params} 
        & \textbf{Memory (GB)} & \textbf{Params} \\
        \midrule
        DGI    & 3.73  & $7.3\times10^5$  & 3.85  & $1.9\times10^6$  & 3.66  & $2.6\times10^5$  \\
        GMI    & 4.06  & $9.9\times10^5$  & 4.20  & $2.2\times10^6$  & 3.93  & $5.2\times10^5$  \\
        MVGRL  & 2.25  & $9.9\times10^5$  & 2.55  & $2.2\times10^6$  & 2.37  & $5.2\times10^5$  \\
        SUGRL  & 1.57  & $9.7\times10^5$  & 1.71  & $2.6\times10^6$  & 1.70  & $3.9\times10^5$  \\
        S$^3$-CL & 1.37  & $7.3\times10^5$  & 1.62  & $1.9\times10^6$  & 1.54  & $2.6\times10^5$  \\
        \midrule
        Ours   &  0.70    &$3.5\times10^5$               & 0.80      &$6.4\times10^5$                  &      1.90 &$2.3\times10^5$                  \\
        \bottomrule
    \end{tabular}
    \label{tab:efficiency}
\end{table*}

\subsubsection{Study on Performance against Test Time Node Feature Distortion}
We evaluate the performance of our method on node classification by augmenting the test splits. We sample $p$ percentage test nodes randomly and replace their node features with points sampled from a standard Gaussian distribution. We evaluate the performance of the distorted test nodes on Amazon Computers, WikiCS, Amazon Photos and Coauthor CS by varying $p$ between 10\% and 40\% and report their performance differences in table \ref{tab:distortion}. In this study, we only use WikiCS, Co-author and Amazon networks since Planetoid datasets tend to be unstable on evaluation as indicated in \citep{shchur2018pitfalls}. Despite no further fine-tuning on distorted node features, the average performance drop on Amazon Photos and Computers is only 2.96\% and 4.01\% respectively. This indicates the ability of the joint framework to generalize on noise-augmented views as well. According to our hypothesis, we believe this is due to the nature of joint predictive embedding learning where a higher variation in context and target embeddings is required for better generalization. 

\begin{table}[h!]
    \centering
    \caption{\textbf{Performance on distorting node features at test time.} Values represent the percentage decrease in the model’s performance compared to its original score. A lower value indicates better performance.}
    \begin{tabular}{lcccc}
        \toprule
        \textbf{Ratio} & \textbf{Photos} & \textbf{Computers} & \textbf{Coauthor CS} & \textbf{WikiCS} \\
        \midrule
        0.10 & -3.3\% & -3.8\% & -3.3\% &-8.9\%  \\
        0.15 & -0.8\% & -3.4\% & -1.2\% &-9.7\%  \\
        0.20 & -3.7\% & -3.3\% & -3.3\% & -13.6\% \\
        0.25 & -2.9\% & -4.6\% & -7.7\% &-12.6\%  \\
        0.30 & -3.9\% & -4.3\% & -5.9\% & -13.2\%\\
        0.35 & -2.4\% & -4.0\% & -6.6\% & -12.3\%\\
        0.40 & -3.7\% & -4.7\% & -9.6\% & -15.2\% \\
        \bottomrule
    \end{tabular}
    \label{tab:distortion}
\end{table}

\section{Conclusion}
In this paper, we propose a novel Graph Self-Supervised Learning (Graph-SSL) framework that combines joint predictive embedding and pseudo-labeling to effectively capture global knowledge while avoiding noisy features and bypassing contrastive methods such as negative sampling and reconstruction. The joint predictive embedding framework leverages the context-target relationship between node embeddings in the latent space by predicting multiple target embeddings for a single context. This approach, combined with the optimization of node feature contributions to pseudo-labels, enables a lightweight Graph Neural Network (GNN) encoder to capture intricate patterns in both graph structure and node features without requiring the stacking of multiple layers or encoders. Additionally, our method addresses the node representation collapse problem by incorporating information from multiple targets for a single context, ensuring robust and diverse embeddings. Through extensive experiments on multiple benchmark graph datasets, we demonstrate that our proposed framework achieves superior performance compared to several state-of-the-art graph self-supervised learning methods.

\section{Conclusion}
In this paper, we propose a novel Graph Self-Supervised Learning (Graph-SSL) framework that combines joint predictive embedding and pseudo-labeling to effectively capture global knowledge while avoiding noisy features and bypassing contrastive methods such as negative sampling and reconstruction. The joint predictive embedding framework leverages the context-target relationship between node embeddings in the latent space by predicting multiple target embeddings for a single context. This approach, combined with the optimization of node feature contributions to pseudo-labels, enables a lightweight Graph Neural Network (GNN) encoder to capture intricate patterns in both graph structure and node features without requiring the stacking of multiple layers or encoders. Additionally, our method addresses the node representation collapse problem by incorporating information from multiple targets for a single context, ensuring robust and diverse embeddings. Through extensive experiments on multiple benchmark graph datasets, we demonstrate that our proposed framework achieves superior performance compared to several state-of-the-art graph self-supervised learning methods.


\bibliographystyle{unsrt}
\bibliography{references}

\begin{IEEEbiographynophoto}{Srinitish Srinivasan} He is currently a fourth year undergraduate student in Vellore Institute of Technology, Chennai Campus pursuing Computer Science and Engineering with specialization in Artificial Intelligence and Machine learning. He has been involved in research work in agriculture, pattern recognition and bio-informatics using deep learning. His research interests include social network theory, graph machine learning, self supervised learning. 
\end{IEEEbiographynophoto}

\begin{IEEEbiographynophoto}{Omkumar CU} is currently serving as an Assistant Professor (Sr. G) at the School of Computer Science Engineering, Vellore Institute of Technology Chennai Campus, has a distinguished academic and professional journey. Specializing in Cyber Security, he completed his Ph.D. at Anna University, Chennai, from 2016 to 2020. His educational background includes a B.Tech in CSE (2006-10) and an M.Tech in CSE (2011-2013), both earned with First class with Distinction from institutions affiliated with JNTU-Anantapur. Dr. Kumar embarked on his teaching career at SRM-Easwari Engineering College, Chennai, from 2013 to 2016, where he discovered his passion for teaching and honed his pedagogical skills. His research interests focus on IoT and Deep Learning, leading to the filing of an International Patent and the publication of 25 research papers in reputable journals, establishing an H-Index value of 6. 
\end{IEEEbiographynophoto}

\vfill

\end{document}